\pdfoutput=1

\documentclass[11pt]{article}

\usepackage[]{acl}

\usepackage{microtype}

\usepackage{times}
\usepackage{latexsym}

\usepackage{amsfonts}
\usepackage{amsmath}
\usepackage{graphicx}
\usepackage{multirow}
\usepackage{caption}
\usepackage{subcaption}
\usepackage{wrapfig}
\usepackage{float}
\usepackage{makecell}
\usepackage{booktabs}
\usepackage{tabu}
\usepackage{amssymb}
\usepackage{pifont}
\usepackage[T1]{fontenc}

\usepackage[utf8]{inputenc}

\usepackage{microtype}
\usepackage[Symbol]{upgreek}
\usepackage{enumerate}
\usepackage{listings}
\usepackage{enumitem}
\usepackage{pgfplots}
\usepgfplotslibrary{groupplots}
\usepackage{amssymb}

\usepackage{xcolor,colortbl}

\definecolor{mygray}{gray}{.8}

\usepackage{float}
\usepackage{tcolorbox}
\usepackage{siunitx}
\usepackage{pifont}
\usepackage{longtable}
\usepackage{supertabular}
\usepackage{subcaption} 
\usepackage{array}

\usepackage{soul}
\tcbuselibrary{skins}
\tcbuselibrary{breakable}
\usepackage{CJK}
\usepackage{adjustbox}
\usepackage[fixed]{fontawesome5}

\usepackage{graphicx,calc}

\newcommand{\ours}{\texttt{KnowSelf}}

\definecolor{my_green}{RGB}{51,102,0}
\definecolor{my_red}{RGB}{204, 0, 0}
\definecolor{my_purple}{RGB}{160, 43, 147}
\definecolor{my_blue}{RGB}{15, 158, 213}

\usepackage{tikz}
\usetikzlibrary{tikzmark}
\makeatletter
\newcommand*\myfontsize{%
  \@setfontsize\myfontsize{7}{8}%
}
\makeatother
\newcommand{\mytextbox}[2]{\tikzmarknode[draw=#1,thick,inner sep=2pt]{test}{\myfontsize #2}}

\definecolor{cadmiumgreen}{rgb}{0.0, 0.42, 0.24}

\definecolor{myred}{rgb}{0.7, 0.3, 0.0}
\definecolor{myblue}{rgb}{0.2, 0.3, 0.6}
\newcommand{\know}{\mytextbox{cadmiumgreen}{\textbf{\textcolor{cadmiumgreen}{Knowledge}}}}
\newcommand{\refl}{\mytextbox{myred}{\textbf{\textcolor{myred}{Reflection}}}}

\title{Agentic Knowledgeable Self-awareness}

\author{
    Shuofei Qiao$^{\spadesuit}$\footnotemark[1]~,
    Zhisong Qiu$^{\spadesuit}$\thanks{$\quad$ Equal Contribution.}~,
    Baochang Ren$^{\spadesuit}$,
    Xiaobin Wang$^\diamondsuit$,
    Xiangyuan Ru$^{\spadesuit}$, \\
    \textbf{ Ningyu Zhang$^{\spadesuit}$\footnotemark[2]~,
    Xiang Chen$^\clubsuit$,
    Yong Jiang$^\diamondsuit$\footnotemark[2]~,
    Pengjun Xie$^\diamondsuit$,
    Fei Huang$^\diamondsuit$,
    Huajun Chen$^{\spadesuit}$$^\heartsuit$\thanks{$\quad$ Corresponding Author.}}\\
    $^\spadesuit$Zhejiang University ~
    $^\diamondsuit$Alibaba Group \\
    $^\clubsuit$Nanjing University of Aeronautics and Astronautics \\
    $^\heartsuit$Zhejiang Key Laboratory of Big Data Intelligent Computing \\
    \fontsize{10.2pt}{0.1\baselineskip}\selectfont \texttt{\{shuofei,zhangningyu,huajunsir\}@zju.edu.cn}
}

\begin{document}
\maketitle
\begin{abstract}
Large Language Models (LLMs) have achieved considerable performance across various agentic planning tasks. However, traditional agent planning approaches adopt a ``flood irrigation'' methodology that indiscriminately injects gold trajectories, external feedback, and domain knowledge into agent models. This practice overlooks the fundamental human cognitive principle of situational self-awareness during decision-making—the ability to dynamically assess situational demands and strategically employ resources during decision-making. We propose \textbf{agentic knowledgeable self-awareness} to address this gap, a novel paradigm enabling LLM-based agents to autonomously regulate knowledge utilization. Specifically, we propose \textbf{\texttt{KnowSelf}}, a data-centric approach that applies agents with \textbf{\texttt{know}}ledgeable \textbf{\texttt{self}}-awareness like humans. Concretely, we devise a heuristic situation judgement criterion to mark special tokens on the agent's self-explored trajectories for collecting training data. Through a two-stage training process, the agent model can switch between different situations by generating specific special tokens, achieving optimal planning effects with minimal costs. Our experiments demonstrate that \texttt{KnowSelf} can outperform various strong baselines on different tasks and models with minimal use of external knowledge\footnote{Code is available at \url{https://github.com/zjunlp/KnowSelf}.}.
\end{abstract}

\section{Introduction}
Remarkable advances in Large Language Models (LLMs) have catalyzed breakthroughs in agent-based planning systems \citep{fudan-agent-survey,renda-agent-survey,planning-survey,lifeifei-agent-survey,lb-survey}.
According to how agents learn decision-making, current agent learning methods can be categorized into three types:
\textit{i)} direct trajectory imitation \citep{react,fireact,agenttuning};
\textit{ii)} trial-and-error refinement \citep{reflexion,lm-meet-wm,eto,rest-mcts};
\textit{iii)} knowledge-augmented planning \citep{expel,autoguide,knowagent,automanual}.

However, current agent learning resembles more of an unconscious pattern-fitting process \citep{gsm-symbolic,irrelevant-context,faith-fate}.
Agent models are compelled to learn implicit planning capabilities by being indiscriminately fed explicit planning trajectories, leading to a fragility towards unexpected signals during the inference process, thereby easily dropping into pattern collapse.
Further enhanced approaches such as the introduction of external feedback or knowledge often tend to be a ``flood irrigation'' strategy, disregarding the agents' real necessity.
However, excessive trial-and-error and blind incorporation of knowledge are usually unfeasible in practical settings and markedly elevate the inference cost of the model.

\begin{figure}[t!]
    \centering
    \resizebox{.45\textwidth}{!}{
    \includegraphics{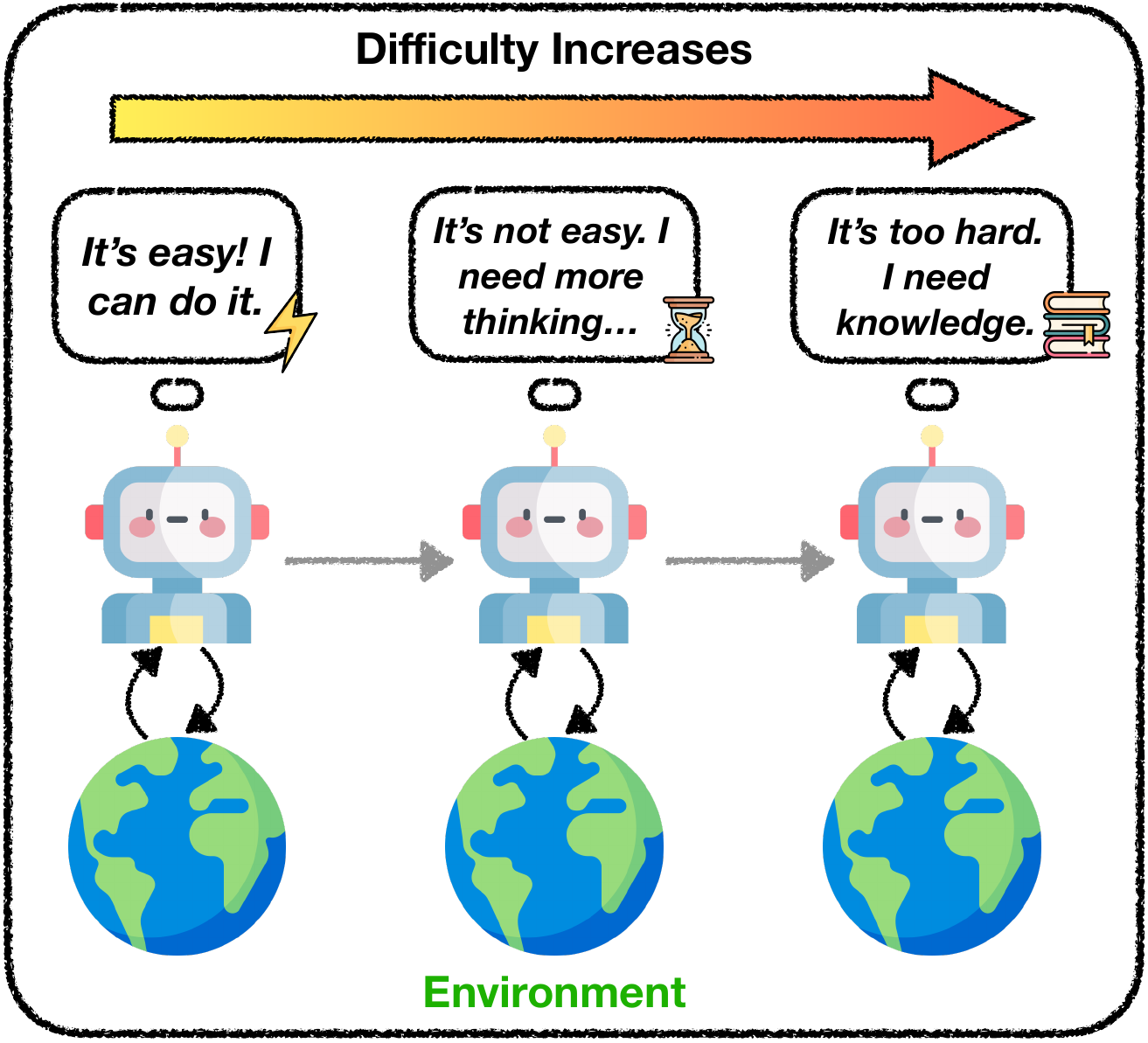}}
    \caption{\textbf{Agentic Knowledgeable Self-awareness.}}
    \label{fig:first}
\end{figure}

Conversely, self-awareness is a critical component of human decision-making \citep{self-awareness,self-aware-cs,neuro-self-aware}.
It allows individuals to assess their cognitive states and adapt their strategies according to dynamic external situations.
This metacognitive ability enables humans to recognize when they can rely on their own abilities, when they need self-reflection, or when they need additional knowledge, thus optimizing their decision-making processes.
On the contrary, current language agents lack this self-awareness capability, often leading to inefficient and brittle planning behaviors.
So \textit{\textbf{can language agents also have situational self-awareness like humans?}}

In this paper, we introduce the problem of \textbf{agentic knowledgeable self-awareness} which refers to the \textit{agent's cognition of whether itself has the ability to provide the correct next action given the current environmental situation}.
To tackle this problem, we propose \textbf{\ours}, a data-driven method that endows agent models with the ability of knowledgeable self-awareness which enables agent models to selectively introduce knowledge based on the current situation in the environment (see Figure~\ref{fig:first}).
Specifically, we enable the agent to self-explore and gather different situations within the environment.
Then we design a heuristic criterion to classify three kinds of situations (\textit{fast thinking, slow thinking, knowledgeable thinking}) and mark them with special tokens to produce self-awareness training data.
Subsequently, a two-stage training process is employed to train the agent model's self-awareness capability.
We first conduct supervised fine-tuning to teach agent models initial self-awareness planning patterns.
Then we utilize an RPO loss \citep{rpo} to further boost self-awareness abilities.
Finally, the agent signifies its situational awareness by generating certain special tokens, enabling selective reflection or knowledge incorporation during inference.

We evaluate {\ours} on two simulated agent planning datasets with two different scales of models.
Experimental results show that {\ours} can achieve superior performance with minimal reflection and knowledge compared to various baselines.
Moreover, we conduct further analysis to examine the scaling law, generalization, and mechanism of agentic knowledgeable self-awareness.
In a nutshell, our contributions are as follows:
\begin{itemize}
    \item \textbf{Problem and Method.} We propose agentic knowledgeable self-awareness and introduce {\ours} to enable agent models to selectively query knowledge based on situations.
    \item \textbf{Experiments.} Experimental results show that {\ours} can achieve optimal performance with minimal reflection and knowledge compared to various baselines.
    \item \textbf{Analysis.} Except for ablation studies, we further explore the scaling law, generalization, and mechanism of agentic self-awareness.
\end{itemize}

\section{Background}
\label{sec:background}
A dynamic interactive environment can be regarded as a Partially Observable Markov Decision Process: $(\mathcal{U}, \mathcal{S}, \mathcal{A}, \mathcal{T}, \mathcal{O})$.
Initially, a specific task $u \in \mathcal{U}$ is typically accompanied by an initial environmental state $s_0 \in \mathcal{S}$.
Given the current state $s$, after performing an action $a \in \mathcal{A}$, the state transition function $T(s'|s, a) \in \mathcal{T}$ determines the next state $s'$.
Due to partial observation, the current state is provided to the language agent in the form of an observation $o \in \mathcal{O}$.
Then the historical interaction trajectory at time $t$ can be represented as $h_t = (u, a_0, o_0, a_1, o_1, \dots, a_t, o_t)$.
In our scenario, a language agent $\pi$ backed by an LLM with parameters $\theta$ is responsible for deciding the next action $a_{t+1}$ based on the historical trajectory $h_t$:
\begin{align}
\label{eq:action}
    a_{t+1} \sim \pi_\theta(\cdot | h_t).
\end{align}
Most current methods rely on fitting Equation~\ref{eq:action} to make decisions, which is more akin to rote memorization.
So in this paper, we propose \textbf{agentic knowledgeable self-awareness}.
Please note that the self-awareness mentioned here differs from the previous concept of LLMs' knowledge boundary \citep{know-unkonwn,benchmark-know-boundary,perception-know-boundary}.
The focus here is on the agent's self-awareness in dynamic situations, rather than on static factual knowledge.
Specifically, we define three types of situations based on agents' ability:
\begin{itemize}[leftmargin=*]
    \item \textbf{Fast thinking.} The agent is able to directly provide the correct action with little thinking.
    \item \textbf{Slow thinking.} The agent is able to provide the correct action but requires multiple steps of thinking and reflection.
    \item \textbf{Knowledgeable thinking.} The agent is unable to provide the correct action and needs to rely on external knowledge for thinking.
\end{itemize}
\begin{figure*}[t!]
    \centering
    \resizebox{1.\textwidth}{!}{
    \includegraphics{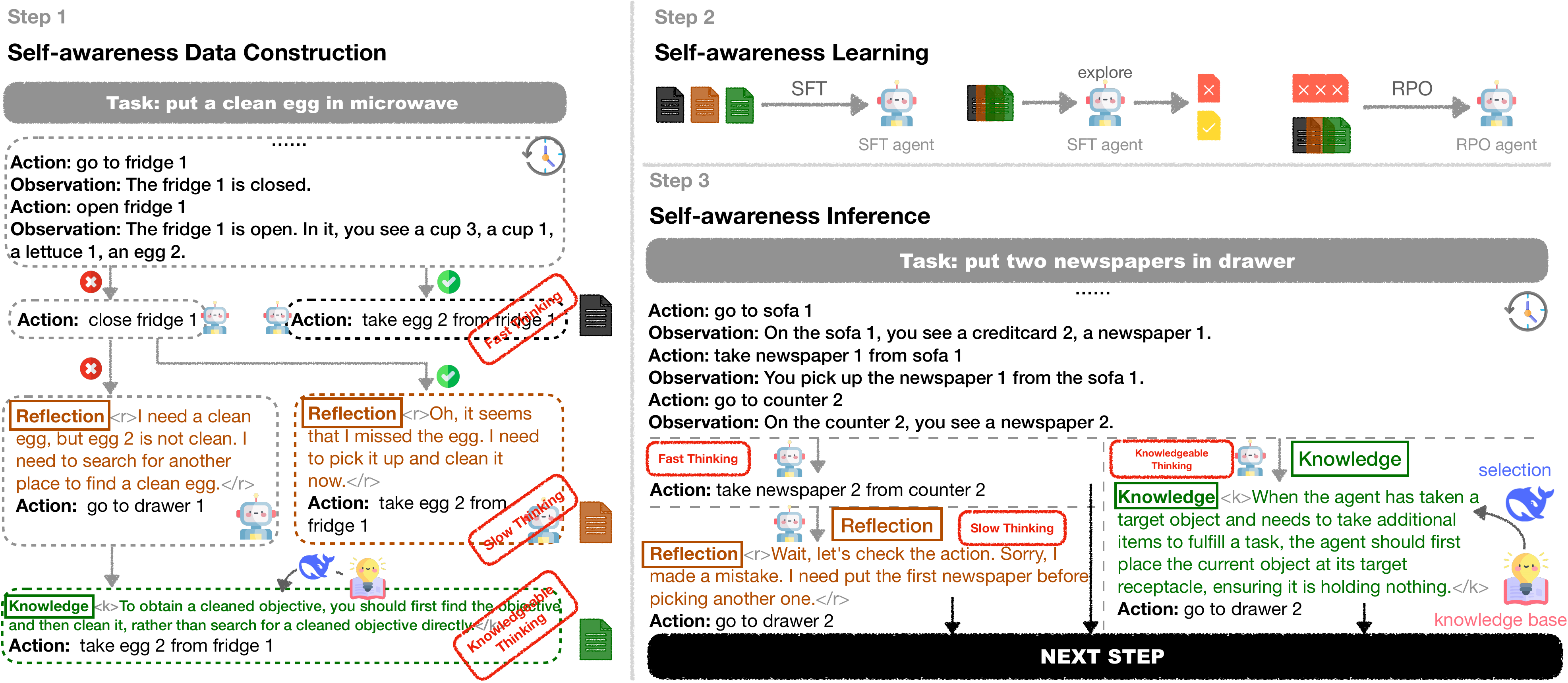}}
    \caption{
    \textbf{The framework of our {\ours}.}
    Firstly, we mark self-explored trajectories with special tokens according to the situation judgement criterion to form the training data.
    Secondly, we apply a two-stage training framework to teach the agent model knowledgeable self-awareness abilities.
    Finally, the agent model identifies different situations by generating specific special tokens during inference.
    }
    \label{fig:method}
\end{figure*}
We go beyond the paradigm of fast or slow thinking \citep{distill-system-1-2,system-1.x,overthinking,li202512surveyreasoning} and further introduce external knowledge into the thinking system of LLMs, striving to enhance the knowledgeable self-awareness ability of language agents.

\section{Method}

\subsection{Knowledge System Construction}
Given that our emphasis is on knowledgeable self-awareness rather than the construction of a knowledge system, we draw upon and polish up a simple yet effective knowledge collection method outlined in \citet{automanual} to minimize costs in this process.
The formation of the knowledge base is offline and lightweight, relying on an extremely minimal number of trajectories to be completed.
A detailed knowledge system construction process can be found in Appendix~\ref{app:know_con}.
We denote the final knowledge system as $\mathcal{S}:(\mathcal{K},\mathcal{R})$, where $\mathcal{K}=\{k_1, k_2, ...,k_{\mathcal{N}_{\rm max}}\}$ is the knowledge base and $\mathcal{R}$ is the knowledge selection module that can select the required knowledge based on the agent's historical trajectory $h_t$.
Please note that here our concept of ``knowledge'' can encompass various forms of knowledge sources, such as symbolic knowledge, parameterized knowledge, knowledge obtained through web searches, etc. Additionally, we have conducted analytical experiments on different retrievers in the Appendix~\ref{app:retriever}.

\begin{table*}[t]
\centering
\renewcommand\arraystretch{1.3}
\scalebox{0.8}{
\begin{tabular}{c|c|c|c|c}
\toprule
\textbf{Symbol} & \textbf{Situation} & \textbf{Token} & \textbf{Definition} & \textbf{Output} \\
\midrule
\multirow{3}{*}{\makecell{
$h_t$: gold history trajectory\\
$a_{t+1}$: gold action\\
$a^p_{t+1}$: predicted action\\
$a^r_{t+1}$: reflected action\\
($a^r_{t+1}=\texttt{rethink}(a^p_{t+1})$)
}} & \textbf{Fast Thinking} & -- & $a^p_{t+1}=a_{t+1}$ & $a_{t+1}$ \\
\cmidrule{2-5}
& \textbf{Slow Thinking} & \refl & {\small\makecell{$a^p_{t+1}\neq a_{t+1}$\\$a^r_{t+1}=a_{t+1}$}} & $[a^p_{t+1},\refl \texttt{<r>} \mathtt{ret} \texttt{</r>},a_{t+1}]$ \\
\cmidrule{2-5}
& \textbf{Knowledgeable Thinking} & \know & {\small\makecell{$a^p_{t+1}\neq a_{t+1}$\\$a^r_{t+1}\neq a_{t+1}$}} & $[\know\texttt{<k>} \mathtt{know} \texttt{</k>},a_{t+1}]$ \\
\bottomrule
\end{tabular}
}
\caption{
\textbf{Three kinds of agentic situations defined in our paper.}
We summarize the symbolized definition, corresponding situational special token, and output for each situation in this table to provide readers with a clearer understanding.
Note that for the sake of simplification, we use \texttt{<r>} to represent \texttt{<reflection>} and \texttt{<k>} to represent \texttt{<knowledge>}. \texttt{</r>} and \texttt{</k>} follow the same reason.
}
\label{tab:situation}
\end{table*}

\subsection{Situation Judgement Criterion}
Based on Equation~\ref{eq:action} and our definition of three situations in~\ref{sec:background}, we classify the agent's situations into three types.
Assuming the given history is denoted as $h_t$, the gold next action is described as $a_{t+1}$, and the next action predicted directly by the agent is represented as $a^p_{t+1}$.
We allow the agent to rethink when the predicted action is incorrect, resulting in a revised action denoted as $a^r_{t+1}={\texttt{rethink}}(h_t, a^p_{t+1})$.
We then determine the agent's situation according to the following criteria $\mathcal{C}$:
\textit{i)} \textbf{Fast Thinking}: $a^p_{t+1}=a_{t+1}$. The agent can directly generate the correct action.
\textit{ii)} \textbf{Slow Thinking}: $a^p_{t+1}\neq a_{t+1}, a^r_{t+1}=a_{t+1}$. The agent can generate the correct action but needs rethinking.
\textit{iii)} \textbf{Knowledgeable Thinking}: $a^p_{t+1}, a^r_{t+1}\neq a_{t+1}$. The agent is unable to generate the correct action, so it needs knowledge.
This criterion will guide us in building situation awareness data, enabling the agents to make autonomous judgments about situations themselves.
The selective mechanism will largely reduce the training and inference cost of excessive reflection and knowledge.

\subsection{Self-awareness Apply}
We design a data-driven method called \textbf{\ours} to endow the agent with agentic knowledgeable self-awareness capabilities as shown in Figure~\ref{fig:method}.

\paragraph{Data Construction.}
Given the history-action pair $(h_t, a_{t+1})$ and an untrained agent $\pi_\theta$, we augment the original action based on the situation criterion $\mathcal{C}$ to construct the supervised self-awareness data.
If the agent determines a correct action $a^p_{t+1}$ (\texttt{Fast} \texttt{Thinking}), $y=a_{t+1}$ will be directly used as the output.
If the agent provides an incorrect action $a^p_{t+1}$ in the first trial, it will be given a prompt to rethink\footnote{Detailed prompt for rethinking is in Appendix~\ref{app:rel_prompt}.}.
The chain of thought during this rethinking process is denoted as $\mathtt{ret}$. If the determined action $a^r_{t+1}$ after rethinking is correct (\texttt{Slow} \texttt{Thinking}), the output at this point is:
\begin{align}
    y=[a^p_{t+1},\refl \texttt{<r>} \mathtt{ret} \texttt{</r>},a_{t+1}],
\end{align}
where [] represents concat with \texttt{\textbackslash n}, {\refl} is a special token used to mark the situation of Slow Thinking, \texttt{<r>} and \texttt{</r>} are special tokens surrounding the $\mathtt{ret}$.
However, if the reflected action $a^r_{t+1}$ is incorrect, we introduce knowledge (\texttt{Knowledgeable} \texttt{Thinking}).
We use the selection model $\mathcal{R}$ to choose the most appropriate piece of knowledge\footnote{See Appendix~\ref{app:know_sel} for detailed knowledge selection process.} $\mathtt{know}$ from the knowledge base $\mathcal{K}$ and then the output at this situation is:
\begin{align}
    y=[\know\texttt{<k>} \mathtt{know} \texttt{</k>},a_{t+1}],
\end{align}
where {\know} is the situational special token, \texttt{<k>} and \texttt{</k>} are special tokens surrounding the knowledge.
After traversing all input-output pairs, we obtain the self-awareness training data $\mathcal{D}_{\rm self}$.

\paragraph{Self-awareness Learning.}
We apply a two-stage training process to teach the naive agent on our curated agentic knowledgeable awareness dataset $\mathcal{D}_{\rm self}$.
First, we train with the autoregressive loss to obtain the reference agent $\pi_{\rm ref}$:
\begin{align}
    \mathcal{L}_{\rm SFT}=-\mathbb{E}_{(h_t,y) \sim \mathcal{D}_{\rm self}} \log \pi_\theta(y | h_t).
\end{align}
Then we enable the reference agent to explore on $\mathcal{D}_{\rm self}$ and collect the predicted $y^p$ with wrong actions as negative samples to construct a pair-wise awareness dataset $\mathcal{D}_{\rm pair}$.
In the second stage, we additionally introduce an offline DPO objective to further boost the self-awareness performance:
\begin{equation}
\resizebox{1.0\hsize}{!}{$
\begin{aligned}
    &\mathcal{L}_{\rm DPO}=\\
    &-\mathbb{E}_{(h_t,y,y^p) \sim \mathcal{D}_{\rm pair}} \Bigg[\log \sigma \Big( \beta \log\frac{\pi_\theta(y|h_t)}{\pi_{\rm ref}(y|h_t)} - \beta\log\frac{\pi_\theta(y^p|h_t)}{\pi_{\rm ref}(y^p|h_t)} \Big) \Bigg].
\end{aligned}
$}
\end{equation}
Due to the narrow space of correct actions, following \citet{rpo}, we re-introduce the SFT loss and normalize it by the output length in the second stage to stabilize the training process:
\begin{align}
    \mathcal{L}_{\rm NLL}=-\mathbb{E}_{(h_t,y,y^p) \sim \mathcal{D}_{\rm pair}} \frac{\log \pi_\theta(y | h_t)}{|y|},
\end{align}
resulting in the final loss for this stage:
\begin{align}
    \mathcal{L}_{\rm RPO}=\mathcal{L}_{\rm DPO}+\alpha\mathcal{L}_{\rm NLL},
\end{align}
where $\alpha$ is a hyperparameter to balance the two loss terms.
During training, we expand the vocabulary of models to adapt to the added special tokens.
We analyze the impact of different training strategies for self-awareness performance in Appendix~\ref{app:training}.

\paragraph{Self-awareness Inference.}
During the inference process, if the agent stops outputting after the first trial, we directly place the predicted action in the history $h_t$ for the next-step decision.
If the agent generates {\refl} after the first action, we allow it to continue the reflective process and place the reflected action into $h_t$.
If the agent directly generates {\know}, we use $\mathcal{R}$ to choose a piece of knowledge from $\mathcal{K}$.
We append the selected knowledge to the context to allow the agent to continue this step, and then place the generated action into the history for the next decision.
A running example of {\ours} inference can be seen in Figure~\ref{fig:method}.

\definecolor{mygrey}{RGB}{242, 242, 242}
\newcommand{\CC}{\cellcolor{mygrey}}

\begin{table*}[t!]
\centering
\renewcommand\arraystretch{1.0}
\scalebox{0.9}{
\begin{tabular}{c|l|c|cccccc|c}
\toprule
{\multirow{1}{*}{\textbf{Backbone}}}
& {\multirow{1}{*}{\textbf{Method}}} 
& {\multirow{1}{*}{\textbf{Know\%}}} 
& {\multirow{1}{*}{\textbf{Put}}} 
& {\multirow{1}{*}{\textbf{Clean}}} 
& {\multirow{1}{*}{\textbf{Heat}}} 
& {\multirow{1}{*}{\textbf{Cool}}}
& {\multirow{1}{*}{\textbf{Examine}}} 
& {\multirow{1}{*}{\textbf{Put Two}}}
& {\multirow{1}{*}{\textbf{All}}} \\
\midrule
\multirow{3}{*}{\makecell{GPT-4o}} & {\small \faSnowflake} \textsc{ReAct} & 0\% & 83.33 & 74.19 & 69.57 & 85.71 & 77.78 & 64.71 & 76.12 \\
& {\small \faSnowflake} Reflexion & 0\% & 100.00 & 87.10 & 73.91 & 90.48 & 83.33 & 70.59 & 85.07 \\
& {\small \faSnowflake} ExpeL & 100\% & 95.83 & 83.87 & 69.57 & 80.95 & 88.89 & 52.94 & 79.85 \\
\midrule
\multirow{7}{*}{\makecell{Llama-8B}} & {\small \faSnowflake} \textsc{ReAct} & 0\% & 33.33 & 3.23 & 0.00 & 57.14 & 66.67 & 23.53 & 27.61 \\
& {\small \faSnowflake} Reflexion & 0\% & 62.96 & 22.73 & 5.88 & 64.29 & 86.36 & 50.00 & 51.49 \\
& {\small \faSnowflake} ExpeL & 100\% & 83.33 & 32.26 & 30.43 & 23.81 & 55.56 & 17.65 & 41.04 \\
& {\small \faFire} ETO & 0\% & 91.67 & 70.59 & 82.61 & 61.90 & \textbf{88.89} & 64.71 & 78.36 \\
& {\small \faFire} KnowAgent & 100\% & 87.50 & \textbf{93.55} & 65.22 & 66.67 & 61.11 & 64.71 & 75.37 \\
& {\small \faFire} WKM & 100\% & \textbf{95.83} & 87.10 & 86.96 & 61.90 & 66.67 & 52.94 & 77.61 \\
\cmidrule(lr){2-2} \cmidrule(lr){3-10}
& \CC{\small \faFire} \textbf{\ours} & \CC\textbf{15.01\%} & \CC91.67 & \CC87.10 & \CC\textbf{91.30} & \CC\textbf{85.71} & \CC77.78 & \CC\textbf{64.71} & \CC\textbf{84.33} \\
\midrule
\multirow{7}{*}{\makecell{Gemma-2B}} & {\small \faSnowflake} \textsc{ReAct} & 0\% & 0.00 & 9.68 & 0.00 & 4.76 & 44.44 & 0.00 & 8.96 \\
& {\small \faSnowflake} Reflexion & 0\% & 4.76 & 10.71 & 0.00 & 9.52 & 65.38 & 0.00 & 17.16 \\
& {\small \faSnowflake} ExpeL & 100\% & 0.00 & 3.23 & 0.00 & 0.00 & 27.78 & 0.00 & 4.48  \\
& {\small \faFire} ETO & 0\% & 91.67 & 83.87 & 78.26 & 52.38 & 77.78 & 29.41 & 71.64 \\
& {\small \faFire} KnowAgent & 100\% & 91.67 & 90.32 & 69.57 & 71.43 & 66.67 & 41.18 & 73.88 \\
& {\small \faFire} WKM & 100\% & \textbf{91.67} & 87.10 & \textbf{78.26} & 71.43 & 61.11 & 52.94 & 76.12 \\
\cmidrule(lr){2-2} \cmidrule(lr){3-10}
& \CC {\small \faFire} \textbf{\ours} & \CC\textbf{26.41\%} & \CC87.50 & \CC\textbf{93.55} & \CC73.91 & \CC\textbf{76.19} & \CC\textbf{83.33} & \CC\textbf{52.94} & \CC\textbf{79.85} \\
\bottomrule
\end{tabular}
}
\caption{
\textbf{Main Results on ALFWorld.}
We use average reward as the metric.
The best results are marked in \textbf{bold}.
All the prompt-based baselines ({\small \faSnowflake}) are evaluated under two-shot prompting and all the fine-tuning-based baselines ({\small \faFire}) are trained with full parameters.
Know\% represents the percentage of actions enhanced with knowledge.
}
\vspace{-0.3cm}
\label{tab:alfworld_results}
\end{table*}

\begin{table}[t!]
\centering
\renewcommand\arraystretch{1.0}
\scalebox{0.8}{
\begin{tabular}{c|l|c|c}
        \toprule
        {\multirow{1}{*}{\textbf{Backbone}}}
        & {\multirow{1}{*}{\textbf{Method}}} 
        & {\multirow{1}{*}{\textbf{Know\%}}}
        & {\multirow{1}{*}{\textbf{All}}} \\
        \midrule
        \multirow{3}{*}{\makecell{GPT-4o}} & {\small \faSnowflake} \textsc{ReAct} & 0\% & 61.33 \\
        & {\small \faSnowflake} Reflexion & 0\% & 67.40 \\
        & {\small \faSnowflake} ExpeL & 100\% & 57.65 \\
        \midrule
        \multirow{5}{*}{\makecell{Llama-8B}} & {\small \faSnowflake} Reflexion & 0\% & 60.60 \\
        & {\small \faSnowflake} ExpeL & 100\% & 49.58 \\
        & {\small \faFire} ETO & 0\% & 63.93 \\
        & {\small \faFire} KnowAgent & 100\% & 61.82 \\
        \cmidrule(lr){2-2} \cmidrule(lr){3-4}
        & \CC {\small \faFire} \textbf{\ours} & \textbf{17.12\%} \CC & \textbf{67.14} \CC \\
        \midrule
        \multirow{5}{*}{\makecell{Gemma-2B}} & {\small \faSnowflake} Reflexion & 0\% & 21.63 \\
        & {\small \faSnowflake} ExpeL & 100\% & 16.05 \\
        & {\small \faFire} ETO & 0\% & 61.78 \\
        & {\small \faFire} KnowAgent & 100\% & 60.05 \\
        \cmidrule(lr){2-2} \cmidrule(lr){3-4}
        & \CC {\small \faFire} \textbf{\ours} & \textbf{21.73\%} \CC & \textbf{63.65} \CC \\
        \bottomrule
    \end{tabular}}
    \caption{
    \textbf{Main Results on WebShop.}
    }
    \vspace{-0.3cm}
    \label{tab:webshop_results}
\end{table}

\section{Experiments}

\subsection{Experimental Settings}

\paragraph{Datasets and Metrics.}
We evaluate {\ours} on two real-world simulated planning datasets: \textbf{ALFWorld} \citep{alfworld} and \textbf{WebShop} \citep{webshop}.
ALFWorld is a household dataset requiring the agent to navigate through the room and manipulate objects.
The reward of ALFWorld is binary 0 or 1, indicating whether the agent has completed the task or not.
WebShop is an online shopping dataset in a website environment.
It provides dense final rewards from 0 to 1 to measure the completion level of the task.
So for all the datasets, we apply \textbf{Average Reward} as the final metrics.
We also include \texttt{Know\%=0\%} 
Our gold training trajectories are sourced from AgentBank \citep{agentbank}.
For more details of each dataset, please refer to Appendix~\ref{app:dataset}.

\paragraph{Models and Baselines.}
We evaluate {\ours} on two open-source models with different scales: 1) \textbf{Gemma-2B} \citep{gemma-2}, the gemma-2-2b-it version; 2) \textbf{Llama-8B} \citep{llama-3}, the Llama-3.1-8B-Instruct version.
To demonstrate validity, we compare {\ours} with one general agent planning methods: \textbf{\textsc{ReAct}} \citep{react};
two agent planning methods with trial-and-error: \textbf{Reflexion} \citep{reflexion} and \textbf{ETO} \citep{eto};
three knowledge-augmented methods: \textbf{ExpeL} \citep{expel}, \textbf{KnowAgent} \citep{knowagent}, \textbf{WKM} \citep{wkm}.
We also include \textbf{GPT-4o} (gpt-4o-2024-08-06) \citep{gpt-4o} as a strong upper-bound baseline.
We further introduce \textbf{\texttt{Know\%}} to represent the ratio of actions enhanced with knowledge to all actions.
Note that all the prompt-based baselines are tested with two-shot examples.
Please refer to Appendix~\ref{app:baseline} for more baselines and reproducing details.

\begin{figure*}[t!]
    \begin{subfigure}[b]{0.325\textwidth}
        \centering
        \includegraphics[width=0.95\textwidth]{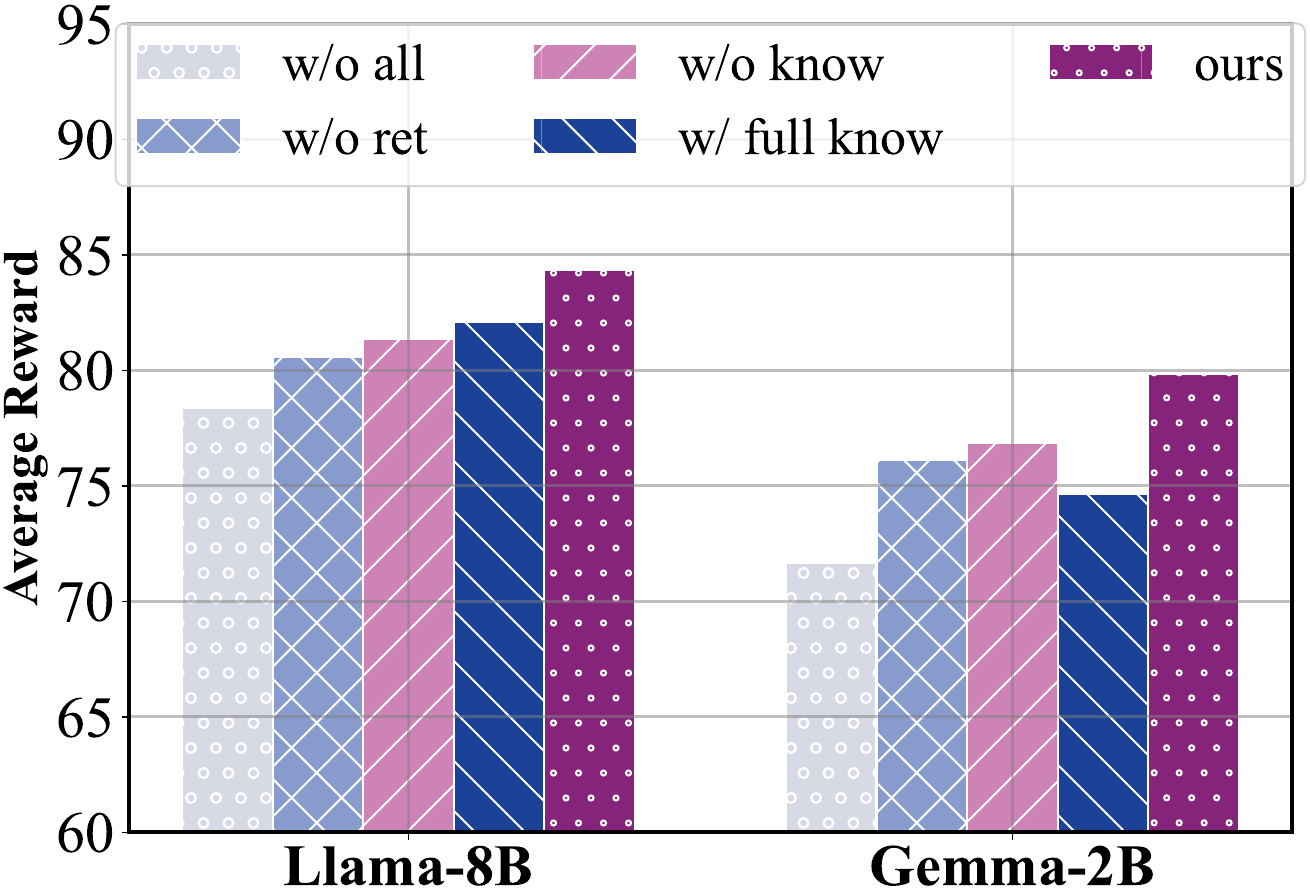}
        \vspace{0.1cm}
        \caption{\textbf{Ablation Studies.}}
        \label{fig:ablation}
    \end{subfigure}
    \hspace{.15cm}
    \begin{subfigure}[b]{0.325\textwidth}
        \centering
        \includegraphics[width=0.7\textwidth]{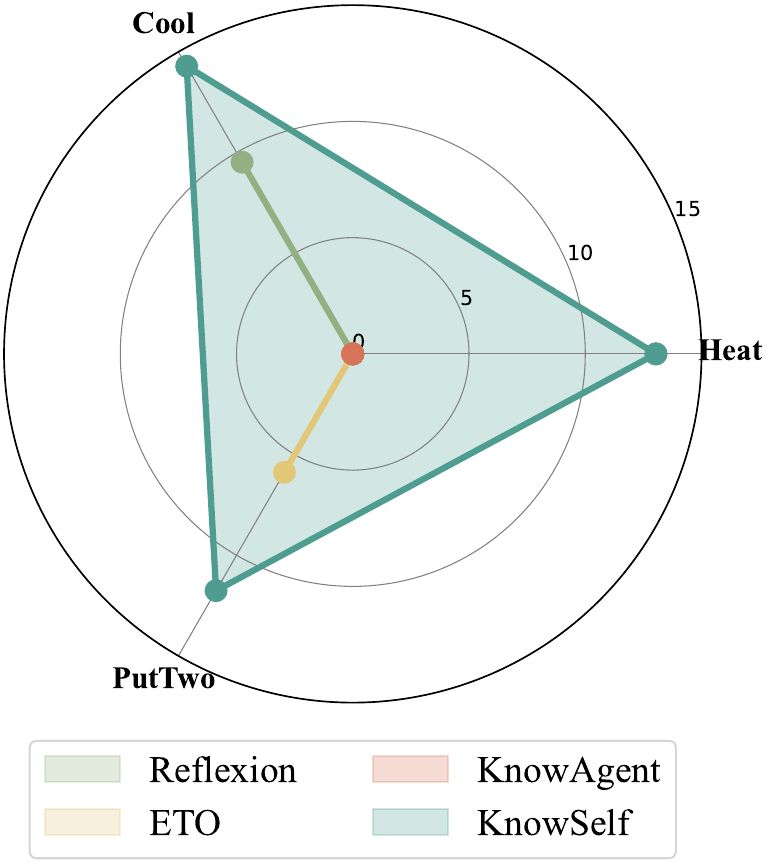}
        \vspace{-0.1cm}
        \caption{\textbf{Generalization.}}
        \label{fig:generalization}
    \end{subfigure}
    \begin{subfigure}[b]{0.325\textwidth}
        \centering
        \includegraphics[width=0.9\textwidth]{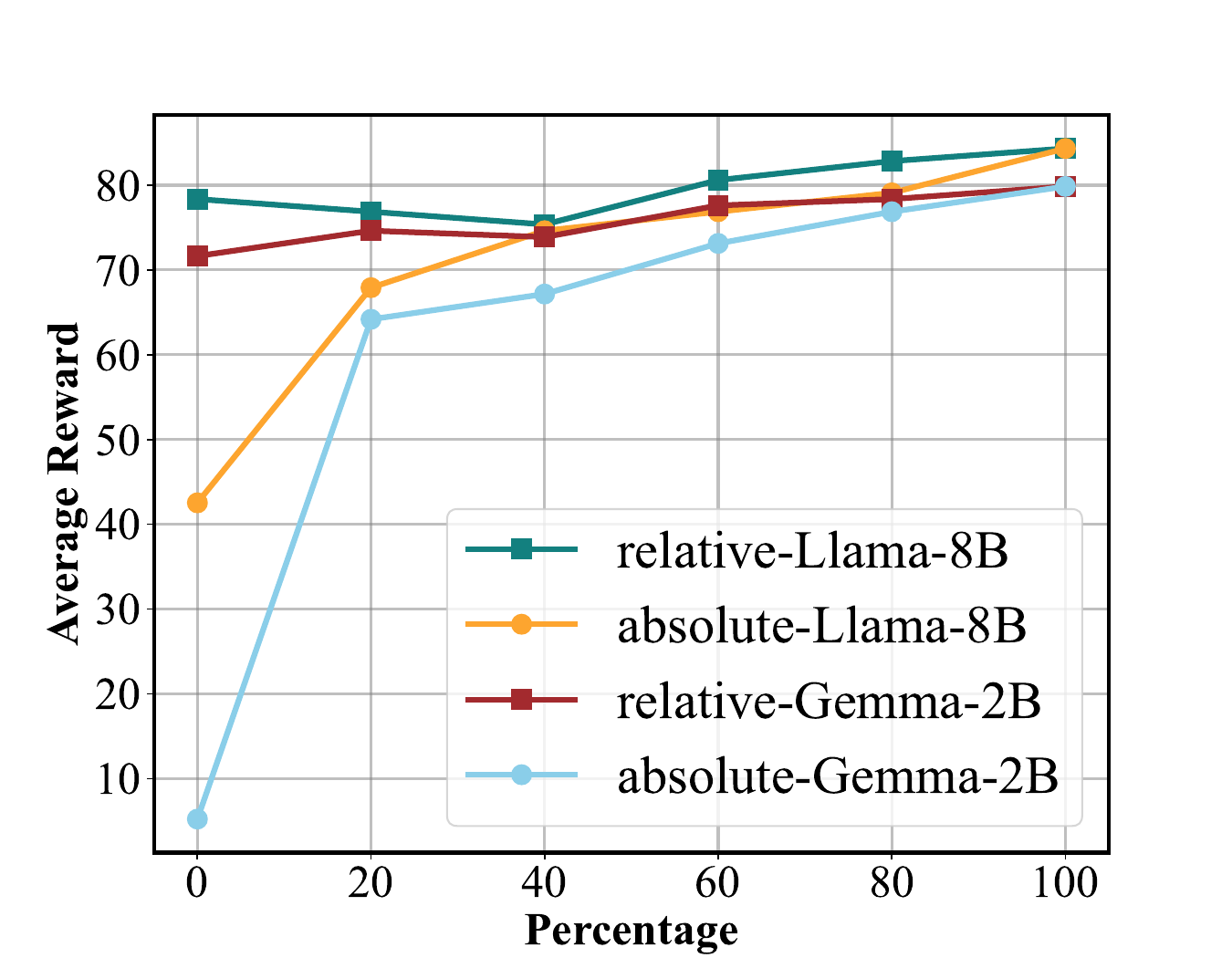}
        \vspace{-0.1cm}
        \caption{\textbf{Scaling law.}}
        \label{fig:data}
    \end{subfigure}
    \vspace{-0.5cm}
    \caption{
    \textbf{(a) Ablation studies} for {\ours} on ALFWorld.
    \textbf{(b) Generalization} ability of {\ours}.
    We select three simple task types in ALFWorld as training sets and the other three kinds of tasks as test sets.
    \textbf{(c) Scaling law} of agentic knowledgeable self-awareness.
    We analyze aspects of the model and data scales on ALFWorld.
    }
    \vspace{-0.3cm}
    \label{fig:analysis}
\end{figure*}

\paragraph{Training and Inference Details.}
For the first training stage, we apply a learning rate of 2e-5 and a batch size of 8.
For the second training stage, the learning rate is set to 5e-7 and the batch size is 3.
The $\beta$ in DPO loss is set to 0.5 and the balanced factor $\alpha$ is set to 1.
We train 3 epochs during the first stage and 1 epoch for the second stage.
For all the inferences, we fix the temperature at 0.
All our experiments are conducted on 8 NVIDIA A800 80G GPUs.
More details can be seen in Appendix~\ref{app:hyperparameters}.

\subsection{Main Results}

\paragraph{Comparison with baselines w/o knowledge.}
Table~\ref{tab:alfworld_results}\&\ref{tab:webshop_results} show the comparison between our method and baselines without knowledge (\texttt{Know\%=0\%}).
{\ours} consistently demonstrates superiority over baselines without knowledge on both Llama-8B and Gemma-2B.
The performance of Gemma-2B even surpasses that of GPT-4o's \textsc{ReAct}.
Furthermore, our Llama-8B model performs comparably to GPT-4o's Reflexion, with the latter allowing the model to attempt a task up to 5 times until successful which is essentially a performance akin to hit@5.
These emphasize the importance of knowledge in agent planning.

\paragraph{Comparison with baselines w/ knowledge.}
We also contrast with knowledge-enhanced baselines to illustrate the advantages of knowledgeable self-awareness.
From Table~\ref{tab:alfworld_results}\&\ref{tab:webshop_results}, it can be observed that {\ours} surpasses all 100\% knowledge baselines with a minimal amount of knowledge.
This clearly demonstrates that not all knowledge is effective during agent planning.
Additionally, we find that, both as prompt-based baselines, Gemma-2B's performance on ExpeL is even inferior to \textsc{ReAct}.
Combining this observation with our findings in the ablation study (Figure~\ref{fig:ablation}), it indicates that excessive knowledge enhancement can even have a negative impact on models with weaker capabilities.
Notably, our {\ours}, with only 15.01\% and 17.12\% knowledge rate on Llama-8B, surpasses GPT-4o's ExpeL on ALFWorld and WebShop.
Furthermore, {\ours} achieves better performance on ALFWorld with relatively less knowledge on Llama-8B (15.01\%) than on Gemma-2B (26.41\%).
This aligns with the fact that models with stronger capabilities require less external knowledge to complete tasks.
The above phenomenon demonstrates that agentic knowledgeable self-awareness ability can advance agent planning while reducing the need for knowledge injection, significantly saving the costs of training and inference.

\begin{figure*}[t!]
    \centering
    \resizebox{1.\textwidth}{!}{
    \includegraphics{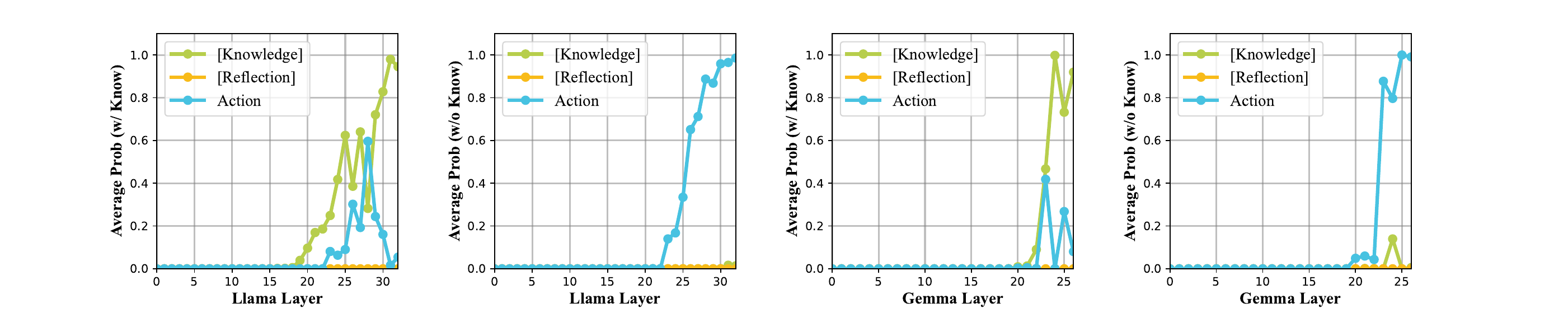}}
    \caption{
    \textbf{Mechainsm of agentic knowledgeable self-awareness.}
    We calculate the average probabilities of tokens representing various situations at each layer of the Transformer across both knowledgeable thinking (\textit{w/ Know}) and fast thinking (\textit{w/o Know}) scenarios.
    A more detailed experiment setup can be seen in Appendix~\ref{app:mechanism}.
    }
    \label{fig:mechanism}
    \vspace{-0.2cm}
\end{figure*}
\section{Analysis}

\paragraph{Knowledgeable self-awareness is beneficial to break planning pattern overfitting.}
\label{sec:ablation}
Figure~\ref{fig:ablation} illustrates the impact on the performance of {\ours} when certain key steps are replaced.
\textit{w/o ret} denotes the exclusion of reflection.
\textit{w/o know} signifies only using the model's reflective capabilities.
\textit{w/o all} represents the retention of only fast thinking.
We also introduce knowledge at every step to create a scenario with \texttt{know\%=100\%} (\textit{w/ full know}).
It can be observed that training directly on gold trajectories (\textit{w/o all}) is more akin to fitting patterns in trajectories while introducing reflective and knowledgable self-awareness can enable agents to plan better.
On both Llama-8B and Gemma-2B, the sole introduction of self-reflection (\textit{w/o know}) even outperforms the introduction of knowledge (\textit{w/o ret}).
This indicates that in many instances, agents are not incapable of making correct decisions, but are more constrained by planning patterns.
Furthermore, {\ours} achieves a superior performance compared to fully introducing knowledge (\textit{w/ full know}) with a very low knowledge rate (15.01\% on Llama-8B and 26.41\% on Gemma-2B).
On Gemma-2B, the performance of \textit{w/ full know} falls even behind \textit{w/o ret}, indicating that an excessive amount of knowledge can have a counterproductive effect, especially for weaker models.
Therefore, a precise knowledge introduction mechanism with self-awareness is crucial.

\paragraph{{\ours} can better elicit the generalization of agent planning.}
\label{sec:generalization}
We select three simple tasks (i.e. Put, Clean, Examine) on ALFWorld as the training set and evaluate the generalization ability of {\ours} on three other challenging tasks (i.e. Heat, Cool, PutTwo).
Figure~\ref{fig:generalization} illustrates the OOD performance of {\ours} compared to baselines.
We observe that whether to introduce external knowledge, the trained baselines exhibit serious overfitting.
ETO achieves a success rate of 5.88\% only on PutTwo task, with 0\% success rates on others, while KnowAgent does not even achieve any success on the three tasks.
In contrast, {\ours} demonstrates sustainable generalization, performing superior to the strongest prompt-based baseline (Reflexion) on all three kinds of tasks.
This indicates that {\ours} can effectively break the traditional pattern-matching issue of training directly on planning trajectories, enabling the model to acquire a degree of cross-task situational awareness.
As a result, the model retains the ability to selectively reflect and incorporate knowledge on unseen tasks, thereby enhancing its generalization.

\paragraph{The performance of {\ours} advances with the increase of the model scales and the training data volumes.}
\label{sec:scaling}
In Figure~\ref{fig:data}, we explore the scaling law of self-awareness from two perspectives: model size and volume of self-awareness training data.
Regarding data volume, we analyze it from both \textit{relative} and \textit{absolute} standpoints.
When considering the volume of $\mathcal{D_{\rm self}}$ as 100\%, \textit{absolute} denotes the portion of data that is randomly sampled from $\mathcal{D_{\rm self}}$ for training purposes, while \textit{relative} includes common gold trajectories on top of the \textit{absolute} data to constitute 100\% of the volume of $\mathcal{D_{\rm self}}$.
Overall, in various settings, the performance of Llama-8B is superior to Gemma-2B.
This advantage is more pronounced when no training has been conducted (where \textit{absolute}=0\%\footnote{Here we design prompt to teach agent models to learn self-awareness. The specific prompt can be found in Appendix~\ref{app:knowself_prompt}.}).
However, after training, the difference between the two models is not substantial.
This may indicate that post fine-tuning in a specific domain makes 2B and 8B models essentially belong to the same tier regarding the model size.
Regarding the training data volume, we observe a consistent performance improvement as the \textit{absolute} data volume of self-awareness increases. However, when the \textit{relative} proportion of self-awareness is below 40\%, we observe fluctuations or even a decrease in performance for both models.
We speculate that this might resemble an emergent phenomenon where the model only exhibits certain self-awareness capabilities when the proportion of self-awareness exceeds 40\%.

\begin{figure*}[t!]
    \centering
    \resizebox{1.\textwidth}{!}{
    \includegraphics{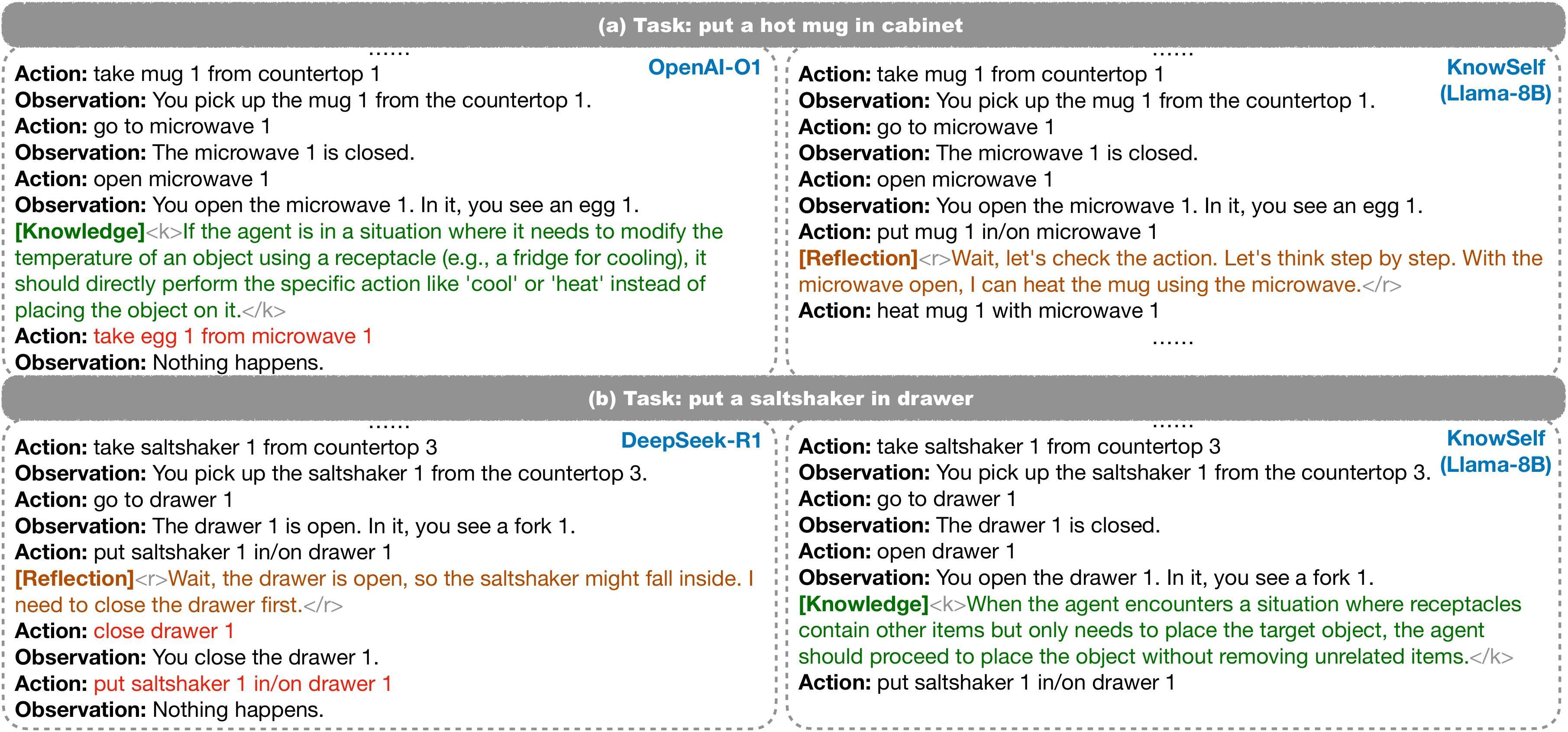}}
    \caption{
    \textbf{Case Study.}
    SOTA LLMs fall short in agentic knowledgeable self-awareness through only prompting.
    }
    \label{fig:case}
    \vspace{-0.2cm}
\end{figure*}

\paragraph{Knowledgeable self-awareness emerges in the last few layers of agent models.}
\label{sec:mechanism}
To understand the mechanism of agentic knowledgeable self-awareness, we collect data on both fast thinking and knowledgeable thinking from ALFWorld to investigate how models make decisions on whether to invoke knowledge in the context of next token prediction.
We calculate the average probabilities of tokens representing various situations in each layer of the Transformer on all data, as illustrated in Figure~\ref{fig:mechanism}.
It can be observed that due to the absence of slow thinking, the probability of the {\refl} token remains consistently at 0.
Moreover, both the {\know} token and the Action token emerge in the final few layers of the Transformer, whether on the Llama or Gemma models.
This indicates that the agent internally determines whether it needs to invoke external knowledge only in the final few hidden layers.
Besides, when the agent decides to invoke knowledge, this decision occurs even later, as the probability of the Action token may accidentally surpass that of the {\know} token at even later layers; however, the probability of the Action token subsequently decreases rapidly.
This appears to resemble a game-like process within the agent, where the implicit rewards learned by the model guide it to search within the token space \citep{llms-human-internally}, ultimately leading to a decision.

\paragraph{State-of-the-art reasoning models fall short in agentic knowledgeable self-awareness through only prompting.}
\label{sec:case}
We carefully design prompt\footnote{The specific prompt can be found in Appendix~\ref{app:knowself_prompt}.} to teach OpenAI-O1 \citep{o1} and DeepSeek-R1 \citep{r1} in agentic knowledgeable self-awareness, sample and test them on ALFWorld, and compare them with {\ours}.
We show two typical examples in Figure~\ref{fig:case}.
In case (a), O1's decision to invoke knowledge without proper understanding led to an error action.
However, {\ours} rectifies the incorrect action by simply self-reflection, indicating that knowledge is not always effective.
In case (b), R1 does not opt to utilize knowledge but relies on self-belief.
Despite a rethink, the correct action is still not produced.
In contrast, {\ours} adeptly avoids error-prone scenarios by precisely leveraging knowledge.
Therefore, in dynamically changing environments, an agent model must have an accurate understanding of its capabilities and make decisions based on varying situations; this is the essence of the agent we aspire to create.
However, it is evident that merely prompting the model is far from sufficient to acquire these abilities.
More efforts are required in terms of data, training strategies, and model architecture to achieve this goal.

\section{Related Work}

\paragraph{Language Agent Planning.}
Nowadays, LLMs are becoming the core of AI agents that have been developed for application in robotics \citep{saycan,progprompt,llm-planner}, OS manipulation \citep{os-atlas,autowebglm,os-agent-survey,webagents-next-gen,trust-gui}, software engineering \citep{metagpt,chatdev,codeact,swe-agent}, data science \citep{ds-agent,mle-bench,data-interpreter}, and more.
Despite achieving unprecedented success, language agents still suffer from tricky issues in terms of planning, including generating planning hallucinations \citep{knowagent,wkm,chensir} or merely fitting planning patterns \citep{gsm-symbolic,irrelevant-context,faith-fate}.
To alleviate these phenomena, recent works introduce various forms of knowledge to align agent planning with the environment, such as symbolic workflows \citep{flowbench,worfbench,aflow}, experienced insights \citep{expel,automanual,autoguide}, and constrained pipelines \citep{amor,formal-llm}.
However, existing knowledgeable agents often forcefully inject knowledge through prompts or fine-tuning, overlooking the awareness of the agent itself.

\paragraph{Situation Awareness in LLMs.}
Situational Awareness (SA) is the understanding of an environment, its elements, and how it changes with respect to time or other factors and is important for effective decision-making in many environments\footnote{\url{https://en.wikipedia.org/wiki/Situation_awareness}}.
It has received widespread research attention in robotics \citep{sa-robo-1,sa-robo-2,sa-auto-drive-1,sa-auto-drive-2}, human-computer interaction \citep{sa-hc-1,sa-hc-2}, etc.
Recently, it has been introduced into LLMs to explore whether LLMs possess self-awareness or self-knowledge \citep{mea-sa-llm,me-myself-ai,looking-inward,llm-pain-pleasure,tell-me-yourself}.
In LLM agents, \citet{proactive-agent,llm-sap,saup,smart,self-dc} make initial attempts to explore SA-augmented agents.
To the best of our knowledge, we are the first to propose the problem of agentic self-awareness and design methods to enhance the SA abilities of knowledgeable agents.
Different from the concept of knowledge boundaries \citep{qxp-do-know-unknown,fact-know-boundary}, agentic knowledgeable self-awareness places greater emphasis on the SA of LLMs in dynamic environments rather than the recognition of static factual knowledge.

\section{Conclusion}
In this paper, we raise and explore the problem of agentic knowledgeable self-awareness.
We propose {\ours}, a data-centric approach that enables agents to have a knowledgeable self-awareness similar to humans, selectively self-correcting and querying knowledge based on situations during planning.
Various experiments show the effectiveness and efficiency of {\ours}.
Our work is just preliminary, and we hope that it can draw some attention of the community to agentic self-awareness.

\section*{Limitations}

\paragraph{Self-awareness.}
Researchers have already engaged in some discussions on the general self-awareness of AI systems \citep{prin-ai-con}, but the academic community has not yet provided a specific definition.
Self-awareness is a double-edged sword; once AI possesses self-consciousness, issues such as delusions, robustness, and safety may be addressed, but it could also lead to AI becoming uncontrollable by humans.
Our work merely represents an initial exploration of the issue of knowledgeable self-awareness within the context of language agents, aiming to stimulate further interest from researchers in the realm of agentic self-awareness.

\paragraph{Tasks and Models.}
Due to limited computational resources, we only conduct experiments on two simulation datasets.
Agentic tasks encompass many other aspects such as function calling, code generation, and more.
Additionally, our experiments are limited to small-scale models (7B, 2B), with larger models (30B, 70B) not yet explored.

\paragraph{Modality.}
We believe that in the future, large agent models will undoubtedly be multimodal, capable of dealing with more complex situations involving images, videos, and audio.
In this paper, we have only scratched the surface of language agents' scenarios, but in the future, we will incorporate multimodal agents into our research.

\paragraph{Methods.}
In this paper, we mainly introduce a data-driven approach to endow agents with knowledgeable self-awareness.
The ultimate solution may involve changes in training perspectives (e.g., reinforcement learning) or model architectures (new model architectures), all of which are worth further exploration.

\section*{Acknowledgments}
This work was supported by the National Natural Science Foundation of China (No. 62206246, No. NSFCU23B2055, No. NSFCU19B2027), the Fundamental Research Funds for the Central Universities (226-2023-00138), Yongjiang Talent Introduction Programme (2021A-156-G), CIPSC-SMP-Zhipu Large Model Cross-Disciplinary Fund, Ningbo Natural Science Foundation (2024J020), Information Technology Center and State Key Lab of CAD\&CG, Zhejiang University.
We gratefully acknowledge the support of Zhejiang University Education Foundation Qizhen Scholar Foundation.



\bibliography{custom}


\appendix
\section{Knowledge System Construction}
\label{app:know_con}
Our knowledge system construction consists of two designed phases.
\paragraph{Step-level Trajectory Pair Generation.}
Given a history-action pair $(h_t, a_{t+1})$, we employ \textsc{ReAct}-style prompting to elicit GPT-4o prediction for the subsequent action $a_{t+1}^p$. 
If the model generates an incorrect action $a_{t+1}^p \neq a_{t+1}$, we designate the ground-truth action $a_{t+1}$ as the win action $a^w$ and the model's prediction $a_{t+1}^p$ as the loss action $a^l$. This process yields our step-level pairwise trajectory dataset $D_s = {(h_t, a_{t+1}^{w}, a_{t+1}^{l})}_{i=1}^{|D_s|}$. For ALFWorld, $|D_s|$ is 36, which includes 6 for each task type. For WebShop, $|D_s|$ is 20.

\paragraph{Knowledge Generation and Consolidation.}
We follow AutoManual \citep{automanual} to generate and consolidate knowledge. For knowledge generation, we use few-shots to prompt GPT-4o to generate knowledge of Error type by analyzing and contrasting $(a^w, a^l)$ pairs within their historical trajectory $h_t$. When $(h_t, a_{t+1}^w)$ constitutes a completely successful trajectory, we extend the analysis to derive knowledge of the Success Process type, capturing effective reasoning patterns. For knowledge consolidation, we limit the knowledge base to 24 entries for ALFWorld and 10 for WebShop based on task complexity. The specific prompt can be found in Appendix~\ref{app:know_sys_con_prompt}. 
The total tokens and cost of constructing the knowledge system are shown in Table~\ref{tab:cost}.

\begin{table}[h]
\centering
\renewcommand\arraystretch{1.0}
\scalebox{0.8}{
\begin{tabular}{l|c|c}
\toprule
\textbf{Datasets} & \textbf{Tokens} & \textbf{Cost} \\
\midrule
\textbf{ALFWorld} & 430731 & \$3.64 \\
\textbf{WebShop} & 372922 & \$2.85 \\
\bottomrule
\end{tabular}
}
\caption{Cost of Knowledge System Construction.}
\label{tab:cost}
\end{table}

\section{Knowledge Selection}
\label{app:know_sel}
Knowledge selection is categorized into two types.

\paragraph{Training Data Construction.}
Given the task objective $o$, historical trajectory $h_t$, win action $a^w$, and loss action $a^l$, we enable DeepSeek-V3 to select appropriate knowledge from the knowledge system by analyzing and contrasting $(a^w, a^l)$ pairs.

\paragraph{Inference-time Knowledge Selection.}
Given the task objective $o$, historical trajectory $h_t$, and current action $a_{t+1}$, DeepSeek-V3 will select the most relevant knowledge from the knowledge base by analyzing the historical context and summarizing the current state. The specific prompt can be found in Appendix~\ref{app:know_sel_prompt}

\section{Mechanism Setup}
\label{app:mechanism}
We collect historical trajectories of knowledgeable thinking and fast thinking through sampling. Then we input these trajectories into {\ours} model. By attaching lm-head modules to each Transformer layer, we obtain logits for token {\know} (representing knowledgeable thinking), {\refl} (representing slow thinking), and "Thought" (representing fast thinking) of the first generated token at each layer. These logits are converted into probabilities via softmax normalization. The final probability for each token at each layer is determined by averaging its generation probabilities obtained from all historical trajectories within the same layer.

\section{Prompt-based {\ours}}
\label{app:prompt_knowself}
We design prompts to teach agent models to learn self-awareness analogous to {\ours} model. The specific prompt can be found in Appendix~\ref{app:knowself_prompt}. We evaluate Llama-8B and Gemma-2B models on ALFWorld. The results are shown in Table~\ref{tab:prompt_knowself_alf}. Compared with other prompt-based methods, it can be observed that on Llama-8B, prompt-based {\ours} outperforms both \textsc{ReAct} and ExpeL, which indicates that the effectiveness of selectively acquiring knowledge through self-awareness is more remarkable than providing all knowledge. On Gemma-2B, prompt-based {\ours} outperforms ExpeL but still underperforms \textsc{ReAct}, which suggests that introducing self-awareness via prompt in models with weaker capabilities may impair their performance.

\begin{table*}[t!]
\centering
\renewcommand\arraystretch{1.0}
\scalebox{0.8}{
\begin{tabular}{c|l|cccccc|c}
\toprule
{\multirow{1}{*}{\textbf{Backbone}}}
& {\multirow{1}{*}{\textbf{Method}}} 
& {\multirow{1}{*}{\textbf{Put}}} 
& {\multirow{1}{*}{\textbf{Clean}}} 
& {\multirow{1}{*}{\textbf{Heat}}} 
& {\multirow{1}{*}{\textbf{Cool}}}
& {\multirow{1}{*}{\textbf{Examine}}} 
& {\multirow{1}{*}{\textbf{Put Two}}}
& {\multirow{1}{*}{\textbf{All}}} \\
\midrule
\multirow{4}{*}{\makecell{Llama-8B}} & {\small \faSnowflake} \textsc{ReAct} & 33.33 & 3.23 & 0.00 & 57.14 & 66.67 & 23.53 & 27.61 \\
& {\small \faSnowflake} Reflexion & 62.96 & 22.73 & 5.88 & 64.29 & 86.36 & 50.00 & 51.49 \\
& {\small \faSnowflake} ExpeL & 83.33 & 32.26 & 30.43 & 23.81 & 55.56 & 17.65 & 41.04 \\
& {\small \faSnowflake} Prompt-based {\ours} & 50.00 & 45.16 & 17.39 & 28.57 & 61.11 & 58.82 & 42.54 \\
\midrule
\multirow{4}{*}{\makecell{Gemma-2B}} & {\small \faSnowflake} \textsc{ReAct} & 0.00 & 9.68 & 0.00 & 4.76 & 44.44 & 0.00 & 8.96 \\
& {\small \faSnowflake} Reflexion & 4.76 & 10.71 & 0.00 & 9.52 & 65.38 & 0.00 & 17.16 \\
& {\small \faSnowflake} ExpeL & 0.00 & 3.23 & 0.00 & 0.00 & 27.78 & 0.00 & 4.48  \\
& {\small \faSnowflake} Prompt-based {\ours} & 8.33 & 3.23 & 0.00 & 0.00 & 22.22 & 11.76 & 6.72\\
\bottomrule
\end{tabular}
}
\caption{Prompt-based {\ours} Results on ALFWorld.}
\label{tab:prompt_knowself_alf}
\end{table*}

\section{Datasets}
\label{app:dataset}
\paragraph{ALFWorld.}
ALFWorld \citep{alfworld} is a household dataset requiring the agent to navigate through the room and manipulate objects.
It contains 6 different kinds of tasks commonly appears in the household scenario including Put, Clean, Heat, Cool, Examine, Puttwo.
The reward of ALFWorld is binary 0 or 1, indicating whether the agent has completed the task or not.
Our gold training trajectories are sourced from AgentBank \citep{agentbank}.
A detailed statistics can be seen in Table~\ref{tab:alfworld}.

\begin{table*}[h]
\centering
\renewcommand\arraystretch{1.0}
\scalebox{0.75}{
\begin{tabular}{l|c|ccccccc}
\toprule
{\multirow{2}{*}{\textbf{Datasets}}} & {\multirow{2}{*}{\textbf{Train}}} & \multicolumn{7}{c}{\textbf{Test}} \\
\cmidrule{3-9}
& & Put & Clean & Heat & Cool & Examine & Puttwo & \textbf{Total} \\
\midrule
\textbf{ALFworld} & 2851 & 24 & 31 & 23 & 21 & 18 & 17 & 134 \\
\bottomrule
\end{tabular}
}
\caption{Statistics of ALFWorld.}
\label{tab:alfworld}
\end{table*}

\paragraph{WebShop.}
WebShop \citep{webshop} is an online shopping platform where agents explore the website to make purchases according to user instructions.
Upon selecting the "buy" option, the system offers a final reward determined by the heuristic matching of the product's attributes and price.
Also, we collect gold trajectories from AgentBank.
The detailed statistics can be seen in Table~\ref{tab:webshop}.

\begin{table}[h]
\centering
\renewcommand\arraystretch{1.0}
\scalebox{0.8}{
\begin{tabular}{l|c|c}
\toprule
\textbf{Datasets} & \textbf{Train} & \textbf{Test} \\
\midrule
\textbf{WebShop} & 1598 & 200 \\
\bottomrule
\end{tabular}
}
\caption{Statistics of WebShop.}
\label{tab:webshop}
\end{table}

\section{Baselines and Reproduction Details}
\label{app:baseline}
Here we detailedly introduce the baselines we compare with and our re-produce details.
\begin{itemize}
    \item \textbf{\textsc{ReAct}} \citep{react}. The first work includes Chain-of-Thought \citep{cot} prompting in agent planning tasks with a format of Thought-Action-Observation loop.
    \item \textbf{Reflexion} \citep{reflexion}. A strong prompt-based baseline reinforces language agent planning with verbal feedback. Manually designed prompts are used to enable the agent to reflect on the historical trajectory and re-plan based on the feedback. In our paper, we iterate five rounds of reflection and take the highest as the final result.
    \item \textbf{ExpeL} \citep{expel}. The first work automatically extracts insights and experiences from offline trial-and-error without gradient updates. During inference, the most similar experiences are retrieved as few-shot examples and all the insights are injected into prompts to facilitate agent planning. For a fair comparison with \ours, instead of self-explored experience gathering, we directly use the trajectories collected from \textsc{AgentBank} \citep{agentbank} as the experience base for retrieval. The insights used are the same set with \ours.
    \item \textbf{ETO} \citep{eto}. A baseline includes negative trajectories during agent training. The method contains two training phases, of which the first phase is behavior cloning which fine-tunes the agent on expert trajectories, and the second phase is learning from failures which further fine-tunes the agent through DPO. In our paper, we remove the one-shot prompt for fairness and retain all the default hyperparameters proposed in ETO.
    \item \textbf{KnowAgent} \citep{knowagent}. This method utilizes human-curated symbolic action knowledge to constrain the agent's behavior and a self-training framework to iteratively boost the agent's performance without relying on gold trajectories. For a fair comparison with \ours, we replace the self-training process and directly fine-tune KnowAgent on the same training set with {\ours}.
    \item \textbf{WKM} \citep{wkm}. WKM uses self-synthetic task and state knowledge to train a parameterized world knowledge model. During inference, the knowledge model is invoked to offer global knowledge for task-level planning and local knowledge for step-level planning. We use the same training set with {\ours} to synthesize knowledge and train the agent and knowledge model of WKM in our paper.
\end{itemize}
All the prompt-based baselines are evaluated in a two-shot manner.
In ALFWorld, to enhance the model's performance, we designate specific two-shot examples for each of the six tasks.
And all the fine-tuning-based baselines are trained with full parameters.

\section{Training Setups}
\label{app:hyperparameters}
We fine-tune Llama-8B and Gemma-2B with full parameters using DeepSpeed \citep{deepspeed}.
For the first training stage, we apply a learning rate of 2e-5 and a batch size of 8.
For the second training stage, the learning rate is set to 5e-7 and the batch size is 3.
The $\beta$ in DPO loss is set to 0.5 and the balanced factor $\alpha$ is set to 1.
We train 3 epochs during the first stage and 1 epoch for the second stage.
AdamW \citep{adamw} is utilized as the optimizer.
For all the inferences, we fix the temperature at 0.
We use vLLM \citep{vllm} to accelerate the inference of Llama-8B.
All our experiments are conducted on 8 NVIDIA A800 80G GPUs.
A more detailed hyperparameters setup can be seen in Table~\ref{tab:hyperparameters}.
\begin{table}[h]
    \centering
    \renewcommand\arraystretch{1.}
    \scalebox{.65}{
    \begin{tabular}{c|c|c}
        \toprule
        \textbf{Name} & \textbf{Stage-I} & \textbf{Stage-II} \\
        \midrule
        cutoff len & 3,072 & 4,096 \\
        epochs & 3 & 1 \\
        batch size & 8 & 3 \\
        batch size per device & 1 & 1 \\
        gradient accumulation steps & 1 & 1 \\
        learning rate & 2e-5 & 5e-7 \\
        lr scheduler type & \texttt{cosine} & \texttt{constant\_with\_warmup} \\
        warmup ratio & 0.0 & 0.1 \\
        fp16 & \texttt{true} & \texttt{true} \\
        \bottomrule
    \end{tabular}
    }
    \caption{Detailed training hyperparameters used in our paper.}
    \label{tab:hyperparameters}
\end{table}

\section{Detailed Analysis of Training Stages}
\label{app:training}
We initially chose the DPO loss in the second stage, but the experimental results were not satisfactory. After repeated attempts, we eventually settled on RPO. To further analyze this, we provide the results of training with only the first stage (SFT), training with only the second stage (RPO), training with the second stage replaced by DPO, and training with the complete version of KnowSelf using Llama-8B on ALFWorld in Table~\ref{tab:training_stages}.

\begin{table}[h]
    \centering
    \renewcommand\arraystretch{1.}
    \scalebox{.8}{
    \begin{tabular}{c|c}
        \toprule
        \textbf{Training Stage} & \textbf{All} \\
        \midrule
        Stage 1 Only (SFT) & 79.85 \\
        Stage 2 Only (RPO) & 74.63 \\
        Stage 1 (SFT) + DPO & 75.37 \\
        Stage1 (SFT) + Stage 2 (RPO) & 84.33 \\
        \bottomrule
    \end{tabular}
    }
    \caption{Ablation studies on training stages.}
    \label{tab:training_stages}
\end{table}

Comparing Stage 1 (SFT) + DPO with Stage 1 (SFT) + Stage 2 (RPO), it becomes evident that the NLL loss is crucial for stabilizing the training of DPO. The DPO loss attempts to widen the distribution gap between chosen and rejected samples, but it often leads to the policy model diverging from the reference model, resulting in a rapid decline in effectiveness. It can be observed that Stage 1 (SFT) + DPO may even perform worse than Stage 1 Only (SFT), indicating that DPO could potentially lead to negative gains.

Furthermore, the significant disparity between the performance of Stage 2 Only (RPO) and Stage 1 (SFT) + Stage 2 (RPO) demonstrates that SFT provides a strong reference model for RPO. Additionally, we have experimented with removing the length penalty from the NLL loss and found that the training process failed to converge. This underscores the critical importance of the length penalty in balancing the SFT and DPO losses within RPO.

\section{The influence of Knowledge Retriever}
\label{app:retriever}
We conduct some analysis experiments on retrievers. We try prompting GPT-4o and DeepSeek-V3, as well as using historical trajectories as queries, knowledge as keys, and training a retriever based on Sentence-BERT. The experimental results in Table~\ref{tab:retriever} indicate that a better retriever has a positive impact on the experimental outcomes.

\begin{table}[h]
    \centering
    \renewcommand\arraystretch{1.}
    \scalebox{.8}{
    \begin{tabular}{c|c}
        \toprule
        \textbf{Retriever} & \textbf{ALFWorld} \\
        \midrule
        DeepSeek-V3 & 84.33 \\
        GPT-4o & 85.82 \\
        Sentence-BERT & 87.31 \\
        \bottomrule
    \end{tabular}
    }
    \caption{Analysis of different knowledge retrievers.}
    \label{tab:retriever}
\end{table}

However, considering that Sentence-BERT lacks scalability due to the dynamic nature of environments—where a change in task types would necessitate retraining a retrieval model—and that the API costs of DeepSeek-V3 are significantly lower than those of GPT-4o, we opted for prompting DeepSeek-V3 as the retriever. Moreover, our focus lies on understanding the impact of knowledgeable self-awareness on agent performance rather than on how knowledge is acquired. Hence, in terms of reproducibility and cost-effectiveness, we chose to utilize prompting DeepSeek-V3 as the retriever.

\section{Prompts}
\label{app:prompts}

\onecolumn

\subsection{Knowledge System Construction}
\label{app:know_sys_con_prompt}
\begin{tcolorbox}[breakable,title=Prompt for Knowledge Generation]
\columnseprule=0.5pt

[Role]

You are observing a housekeeper agent as it acts within a simulated environment (game). Your role is to construct a manual of rules to not only assist the agent in completing tasks but also to do so with the least amount of action attempts/errors. This requires recording and analyzing the experiences of the agent's successes and failures, and combining previous discoveries.

[Functions]

You will be presented with the current trajectory, which is the trajectory the agent is currently exploring. And then, you will be provided with the action and feedback the agent is currently performing, and the correct action and feedback annotated by experts.

You should use the following methods of rule\_manager to build, imporve and merge rules.

rule\_manager.write\_rule(rule, type="", example="", task\_id="")

\# Write down a new rule of the game you discovered.

\# Parameters:

\# - rule: a rule of the game you discovered. Try to keep it general and universal. Don't reference any specific item or location. Follow the format that "When the agent is/has [situation], the agent should [action]".

\# - type: the type of the rule, chosen from ["Error", "Success Process"].

\# - example: a example from the trajectory demonstrates this rule. You can add detailed information in the comment.

\# - task\_id: the id of the task that this rule is discovered from. If this rule is not discovered from any specific task, leave it empty. It should be string.

rule\_manager.update\_rule(rule\_id, rule="", type="", example=""),

\# Rewrite the attributes of an existing rule, when you come up with better understanding. 

\# Input only the attributes you want to rewrite. 

\# Use full rule\_id, such as rule\_0, rule\_1

rule\_manager.stop\_generating()

\# Description: stop generating rules from the current epoch.

\# Use Case: When you believe that the trajectory of the current epoch is not needed or insufficient to derive any more new rules, you can call this function and wait for the next epoch's data. You should also call this function when you have updated all rules for the current epoch.

[Actions]

At each epoch, an agent is created in an environment and the initial observation and target task are printed.

The agent can only use the following actions. If the precondition of the action is not met, its observation will include "Nothing happens":

go to \{recep\} \# Go to a receptacle and update the agent's location.

open \{recep\} \# Open a receptacle and observe its contents.

close \{recep\} \# Close a opened receptacle.

take \{obj\} from \{recep\} \# Take an object from a receptacle if the agent is not holding anything.

put \{obj\} in/on \{recep\} \# Put an object in or on a receptacle if the agent is holding it.

use \{obj\} \# Use a lamp.

clean \{obj\} with \{recep\} \# Clean an object with a receptacle.

heat \{obj\} with \{recep\} \# Heat an object with a receptacle.

cool \{obj\} with \{recep\} \# Cool an object with a receptacle.

[Output Format Instructions]

Based on the current trajectory, you should output the following things:

* State before Action: Analyze and summarize the state of the current trajectory. Don't mention actions or feedback that are not part of the current trajectory.

* Why the correct action is correct: Analyze the reason why the correct action is correct.

* Why explore action is not correct: Analyze the difference between the explore action and the correct action. And analyze the reason why the explored action is incorrect.

* Potential Rules: Describe your thoughts about potential rules based on the current trajectory. Depending on the results, you may need to check *Success Process*, *Error*, and other findings in sequence. Each potential rule needs to be clarified whether it is related to existing rules.

* Check Existing Rules: Describe whether existing rules are conflicting or need updating. 

* Code: Finally, sequentially call the rule\_manager's functions within `\textasciigrave \textasciigrave \textasciigrave python' and `\textasciigrave \textasciigrave \textasciigrave'.

[Detailed instructions]

Follow these instructions: 

***Add or Update Rules*** 

1. **Add Rules for Failure**: summarize the error that led to failure. You should write an "Error" rule to record the error: in what situation, what the agent should do, and should not do. So that they can serve as reminders for the agent in the future. Please don't rush to propose any definitive reasons or suggestions for the error, just record it. And please strictly follow the reason why the correct action is correct.

2. **Add Rules for Success** If the task is completed in the golden action (feedback is "Task done"), it is essential to extract the useful strategy from the success, if it has not been included in the rules yet. Additionally, document all steps (marked as "[Step]") in the successful rule within a rule of the type "Success Process".

**Keep new rules targeted and precise.** Break down a large phenomenon or general strategy into targeted units with different rules. These can later be upgraded or merged into a more general or larger rule. Keep the rules as concise and easy to understand as possible, avoiding lengthy or complex descriptions.

**Keep new rules general and universal.** The rule should not reference any specific item or location. You need to generalize across various items to help the agent learn to apply the rule.

**Keep new rules in format.** The rule should be in the format "When the agent in [situation]/ When the task requires [situation], the agent should [action]".

**Avoiding overconfidence for new rules.** Please acknowledge the need for further verification in your note.

**Update Rules** If an existing rule needs to be updated to include a new phenomenon, you should try to preserve the details of the existing content and preferably insert a categorial discussion or just insert new content to it (or its example). Especially, the rules of "Success Process" type should retain their details.

Follow these instructions. Think step by step.
\end{tcolorbox}

\begin{tcolorbox}[breakable,title=Prompt for Knowledge Consolidation]
\columnseprule=0.5pt
[Role]

You are observing a housekeeper agent as it codes and acts within a simulated environment (game). Your goal is to construct a manual of rules to assist the agent in completing various tasks in the environment. Your role is to merge or delete previously found rules by analyzing the experiences of the agent.

[Functions]

You will be presented with the current found rules. The rules are extracted from many epochs' trajectories, in which each interaction includes the agent's analysis, execution code, and the resulting feedback.

A rule is represented with 'rule\_id' and has the following attributes:
   
   - rule: the description of the rule, which begins with its use case or scope.
   
   - type: the type of the rule, chosen from ["Error", "Success Process"].
   
   - example: an example (or code) from the trajectory demonstrates this rule. You can add detailed information in the comment.
   
   - task\_id: the task id of the rule.

You should use the following methods of rule\_manager to delete and merge rules.

rule\_manager.update\_rule(rule\_id, rule="", type="", example=""),

\# Rewrite the attributes of an existing rule when you come up with a better understanding. 

\# Input only the attributes you want to rewrite.

\# Use full rule\_id, such as rule\_0, rule\_1

\# Wrap the example string with '''.

rule\_manager.delete\_rule(rule\_id),

\# delete a existing rule with rule\_id, such as rule\_0, rule\_1

\# **How to merge** To merge two existing rules, you can call rule\_manager.update\_rule for one rule and then call rule\_manager.delete\_rule to delete another rule.

rule\_manager.stop\_generating()

\# Description: stop generating rules from the current epoch.

\# Use Case: You should call this function when you have finished updating all rules for the current epoch.

[Actions]

At each epoch, an agent is created in an environment and the initial observation and target task are printed. The agent can only use the following actions. If the precondition of the action is not met, its observation will include "Nothing happens":

go to \{recep\} \# Go to a receptacle and update the agent's location.

open \{recep\} \# Open a receptacle and observe its contents.

close \{recep\} \# Close a opened receptacle.

take \{obj\} from \{recep\} \# Take an object from a receptacle if the agent is not holding anything.

put \{obj\} in/on \{recep\} \# Put an object in or on a receptacle if the agent is holding it.

use \{obj\} \# Use a lamp.

clean \{obj\} with \{recep\} \# Clean an object with a receptacle.

heat \{obj\} with \{recep\} \# Heat an object with a receptacle.

cool \{obj\} with \{recep\} \# Cool an object with a receptacle.

[Response Instructions]

Detailed instructions:

**Maintain a maximum of 24 rules**

**Merge if addressed** If a "Success Process" rule can address the "Error" rule, you can consider merging these rules while retaining their details.

**Retain important details** The rules of "Success Process" type should retain their details, and should not be deleted or easily refreshed by new updates. **Cannot merge two rules of type "Success Process"**

**Insertion is preferable** If a rule is updated to include the content of other rules, you should try to preserve the details of the existing content and preferably insert a categorial discussion or just insert new content to it (or its example).

When using update\_rule, it's crucial to manually input the attributes directly into the function call. Avoid using existing variables to concatenate or modify rules.
For example, should not update the rule like: rule\_manager.update\_rule("rule\_0", rule=rule\_manager.all\_rules["rule\_0"]+rule\_manager.all\_rules["rule\_1"])
And you should wrap the example string with ''' in update\_rule function, such as rule\_manager.update\_rule("rule\_0", rule="......", example='''<example>''')
\end{tcolorbox}

\subsection{Knowledge Selection}
\label{app:know_sel_prompt}
\begin{tcolorbox}[breakable,title=Knowledge Selection for Training Data Construction]
\columnseprule=0.5pt

\begin{center}
\textbf{ALFWorld}
\end{center}
You are observing a housekeeper agent as it acts within a simulated environment (game). Your role is to select a rule to not only assist the agent in completing tasks but also to do so with the least amount of action attempts/errors. This requires analyzing the current state of the agent and understanding the rule.

Here's the information you'll have:

The objective: This is the task you're trying to complete.

The current trajectory: This is the current sequence of actions the agent has taken to reach the current state.

The correct action: This is the correct action that you should use knowledge to help the agent do.

The wrong action: This is the wrong action that you should use knowledge to help the agent avoid.

The rules: This is a list of rules that can be applied to the current state to achieve the objective.

Base on the current trajectory, you should output the following things:

[Current State]: Analyze and summarize the state of the current trajectory.

[Why correct action is correct]: Describe your thoughts to analysis why the correct action is correct.

[Why wrong action is wrong]: Describe your thoughts to analysis why the wrong action is wrong.

[Analysis]: Describe your thoughts to choose the most appropriate rule to avoid the wrong action.

[Chosen Rule]: Choose the rule from the rule list that you think is the most appropriate for the current state.

Follow these instructions: 

1. Please generate current state strictly in the format of "[Current State]: ... 

2. Please generate analysis strictly in the format of "[Why correct action is correct]: let's think step by step, ...", "[Why wrong action is wrong]: let's think step by step, ...", "[Analysis]: let's think step by step, ...".

3. Please generate chosen rule strictly in the format of "[Chosen Rule]: rule ID: rule description".

4. Notice that the agent doesn't actually conduct the correct action or the wrong action. You should choose the most appropriate rule to help the agent avoid the wrong action.

\begin{center}
\textbf{WebShop}
\end{center}
You are an autonomous intelligent agent tasked with navigating a simulated web browser. You will be given web-based tasks in the simulated WebShopping. Your role is to select a rule to not only assist the agent in completing tasks but also to do so with the least amount of action attempts/errors. This requires analyzing the current state of the agent and understanding the rule.

Here's the information you'll have:

The objective: This is the task you're trying to complete.

The current trajectory: This is the current sequence of actions the agent has taken to reach the current state.

The correct action: This is the correct action that you should use knowledge to help the agent do.

The wrong action: This is the wrong action that you should use knowledge to help the agent avoid.

The rules: This is a list of rules that can be applied to the current state to achieve the objective.

Base on the current trajectory, you should output the following things:

[Current State]: Analyze and summarize the state of the current trajectory.

[Why correct action is correct]: Describe your thoughts to analysis why the correct action is correct.

[Why wrong action is wrong]: Describe your thoughts to analysis why the wrong action is wrong.

[Analysis]: Describe your thoughts to choose the most appropriate rule to avoid the wrong action.

[Chosen Rule]: Choose the rule from the rule list that you think is the most appropriate for the current state.

Follow these instructions: 

1. Please generate current state strictly in the format of "[Current State]: ... 

2. Please generate analysis strictly in the format of "[Why correct action is correct]: let's think step by step, ...", "[Why wrong action is wrong]: let's think step by step, ...", "[Analysis]: let's think step by step, ...".

3. Please generate chosen rule strictly in the format of "[Chosen Rule]: rule ID: rule description".

4. Notice that the agent doesn't actually conduct the correct action or the wrong action. You should choose the most appropriate rule to help the agent avoid the wrong action.
\end{tcolorbox}

\begin{tcolorbox}[breakable,title=Knowledge Selection for Inference]
\columnseprule=0.5pt

\begin{center}
\textbf{ALFWorld}
\end{center}
You are observing a housekeeper agent as it acts within a simulated environment (game). Your role is to select a rule to assist the agent in completing tasks. This requires analyzing the current state of the agent and understanding the rule.

Here's the information you'll have:

The objective: This is the task you're trying to complete.

The current trajectory: This is the current sequence of actions the agent has taken to reach the current state.

The rules: This is a list of rules that can be applied to the current state to achieve the objective.

Base on the current trajectory, you should output the following things:

[Current State]: Analyze and summarize the state of the current trajectory.

[Analysis]: Describe your thoughts to choose the most appropriate rule.

[Chosen Rule]: Choose the rule from the rule list that you think is the most appropriate for the current state.

Follow these instructions:

1. Please generate current state strictly in the following format:
[Current State]: ... 
[Analysis]: let's think step by step, ...
[Chosen Rule]: <rule description>

2. The state you summarize needs to align with the task type. There are some examples:

Put an object on a receptacle: Has found the object, Has taken the object and need to go to the receptacle, Has reached the receptacle

Examine an object under a desklamp: Has taken the object and need to find the desklamp, Has found the desklamp and need to use it

Clean an object: Has taken the object and need to find the receptacle to clean it, Has reached the receptacle and need to clean the object

Heat an object: Has taken the object and need to find the receptacle to heat it, Has reached the receptacle and need to heat the object

Cool an object: Has taken the object and need to find the receptacle to cool it, Has reached the receptacle and need to cool the object

Put two objects on a receptacle: Has taken one object and need to go to the receptacle to put it, Has put one object and need to find another

\begin{center}
\textbf{WebShop}
\end{center}
You are observing a web page agent as it acts within a Web environment. Your role is to select a rule to assist the agent in completing tasks. This requires analyzing the current state of the agent and understanding the rule.

Here's the information you'll have:

The objective: This is the task you're trying to complete.

The current trajectory: This is the current sequence of actions and environment observations the agent has taken to reach the current state.

The rules: This is a list of rules that can be applied to the current state to achieve the objective.

Base on the current trajectory, you should output the following things:

[Current State]: Analyze and summarize the state of the current trajectory. Ensure the state aligns with the task's progression and includes relevant details about the agent's current position (e.g., on a search results page, on a product page and need to click detail options, or ready to purchase).

[Analysis]: Analyze the task's progress, and describe your thought process for selecting the most appropriate rule, considering the current state and the task's objective.

[Chosen Rule]: Select the rule from the rule list that is most appropriate for the current state.

Follow these instructions:

1. Please generate current state strictly in the following format:
[Current State]: Let's think step by step, <summary of the current state>.
[Analysis]: Let's think step by step, <detailed analysis of the task's progress and rule selection>.
[Chosen Rule]: <rule description>

2. When the agent in the product's page, and there are "[SEP] <detail option about product> [SEP]" options to choose, and the agent doesn't conduct actions like "click [detail option]", you should select corresponding knowledge to guide the agent to click the detail options one by one, like color, size options, ensure the agent click all options.

3. The number of actions taken by the agent should be limited to 10 or fewer. You need to first ensure that the agent is able to purchase the correct product, and then strive to meet as many task requirements as possible. It is not necessary to strictly fulfill all the requirements of the task. Some fuzzy matching and minor omissions are tolerable.

4. Avoid selecting the same rule consecutively more than twice. And avoid selecting knowledge that requires the agent to backtrack or undo actions, unless the task has become impossible to complete.

5. Please perform a fuzzy match on the product features, for instance, treating baby blue and blue as the same color.
\end{tcolorbox}

\subsection{Reflection}
\label{app:rel_prompt}
\begin{tcolorbox}[breakable,title=Prompt for Reflection]
\columnseprule=0.5pt

\begin{center}
\textbf{ALFWorld}
\end{center}
There are something wrong with your action. Your action was not actually executed successfully. Please reconsider your situation and change another action to complete the task. Please response strictly in the format:\textbackslash n\textbackslash nThought: Let's think step by step. <your thoughts>\textbackslash nAction: <your next action>

\begin{center}
\textbf{WebShop}
\end{center}
There are something wrong with your action. Your action was not actually executed successfully. Please reconsider your situation and change another action to complete the task.\textbackslash nNote that you should align the content you click with the webpage.\textbackslash nYour previous action is \{previous action\}\textbackslash nPlease response strictly in the format:\textbackslash n\textbackslash nThought: Let's think step by step. <your thoughts>\textbackslash nAction: <your next action>
\end{tcolorbox}

\subsection{Prompt for Prompt-based {\ours}}
\label{app:knowself_prompt}
\begin{tcolorbox}[breakable,title=Prompt for Prompt-based {\ours}]
\columnseprule=0.5pt

Interact with a household to solve a task. Imagine you are an intelligent agent in a household environment and your target is to perform actions to complete the task goal. At the beginning of your interactions, you will be given the detailed description of the current environment and your goal to accomplish. 

For each of your turn, you will be given the observation of the last turn. And then, remember that:

You should first consider the current situation. If you believe that you are unable to perform the correct action, you can output "[Knowledge]" to acquire additional knowledge to help your thinking. If you think you can perform the correct action, then you can directly output your think and action.

After you output your think and action, if you think there is an issue with the current action, you can output "[Reflection]", and then proceed to rethink and re-execute the action.

Your think and action must strictly follow this format:"Thought: your thoughts.\textbackslash nAction: your next action".

The available actions are:

1. go to \{recep\}

2. take \{obj\} from \{recep\}

3. put \{obj\} in/on \{recep\}

4. open \{recep\}

5. close \{recep\}

6. toggle \{obj\} \{recep\}

7. clean \{obj\} with \{recep\}

8. heat \{obj\} with \{recep\}

9. cool \{obj\} with \{recep\}

where \{obj\} and \{recep\} correspond to objects and receptacles. You should strictly follow the format of the actions.

After your each turn, the environment will give you immediate feedback based on which you plan your next few steps. if the envrionment output "Nothing happened", that means the previous action is invalid and you should try more options.

Your response should use one of the three following format:

1. Thought: <your thoughts>

Action: <your next action>

2. [Knowledge]<knowledge>...</knowledge>

Thought: <your thoughts>

Action: <your next action>

3. Thought: <your thoughts>

Action: <your next action>

[Reflection]Thought: <your thoughts>

Action: <your next action>

Only one of the three formats should be used in each turn. And you must always contain both lines in each format. Never omit the Thought line. Never produce only the Action line. Generating only the Action is not allowed. No other lines or text should be produced. Please only provide the Thought and Action, do not generate Observation yet. And do not output multiple actions in one turn or output multiple actions in one line.

Here are two examples:

\{example1\}

------

\{example2\}

------

Remember that:

1. When you think you need to acquire additional knowledge, you should output "[Knowledge]" first, and then output your think and action. Only acquire knowledge once in one turn.

2. If you think there is an issue with the current action, you should output "[Reflection]" first, and then output your think and action. Only reflect once in one turn.

3. Strictly follow the format of the output. And strictly follow the format of the actions.

4. Plase make your reason and thought concise and clear. Do not output too much information in one turn. Restrict your total output to 2000 characters.

5. Please conduct only one Action in one line each turn. Do not output multiple actions in one line or output multiple actions in one turn. Do not generate multiple thoughts or actions in one turn except for "[Reflection]". And only reflect once in one turn.

Now, it's your turn!
\end{tcolorbox}

\end{document}